# Path Planning with Divergence-Based Distance Functions

Renjie Chen, Craig Gotsman, and Kai Hormann

*Abstract*—Distance functions between points in a domain are sometimes used to automatically plan a gradient-descent path towards a given target point in the domain, avoiding obstacles that may be present. A key requirement from such distance functions is the absence of spurious local minima, which may foil such an approach, and this has led to the common use of harmonic potential functions. Based on the planar Laplace operator, the potential function guarantees the absence of spurious minima, but is well known to be slow to numerically compute and prone to numerical precision issues. To alleviate the first of these problems, we propose a family of novel *divergence distances*. These are based on *f-divergence* of the *Poisson kernel* of the domain. We define the divergence distances and compare them to the harmonic potential function and other related distance functions.

Our first result is theoretical: We show that the family of divergence distances are *equivalent* to the harmonic potential function on simply-connected domains, namely generate paths which are *identical* to those generated by the potential function. For the special case where the domain is a disk, divergence distances are also equivalent to the (Poincaré) hyperbolic metric and generate circular hyperbolic geodesics. The proof is based on the concept of *conformal invariance*.

Our other results are more practical and relate to two special cases of divergence distances, one based on the *Kullback-Leibler* divergence and one based on the *total variation* divergence. To compute all the distance functions, the domain is discretized to a triangle mesh to facilitate a Finite Elements (FEM) computation. We show that using divergence distances instead of the potential function and other distances has a significant computational advantage, as, following a pre-processing stage, they may be computed up to an order of magnitude faster than the others when taking advantage of certain sparsity properties of the Poisson kernel. Furthermore, the computation is "embarrassingly parallel", so may be implemented on a GPU with up to three orders of magnitude speedup.

A disadvantage of using divergence distances is that computing them using very low resolution meshes may introduce spurious local minima into the distances. However, we demonstrate experimentally that this phenomenon quickly disappears at reasonable resolutions.

## I. Introduction

PATH planning in a planar domain containing obstacles is an important problem in robotic navigation. The main challenge is for an autonomous agent to move from one point (the *source*) in the domain to another (the *target*) along a realistic path which avoids the obstacles, where the path is determined automatically and efficiently based only on knowledge of the domain and local information related to the current position of the agent. This important problem has attracted much attention in the robotics community and is the topic of ongoing research. A significant family of path planning algorithms is based on so-called *potential functions*, which are inspired by the physics of electrical force fields, first proposed in the late 1980's by Khatib [14] and developed by Kim and Khosla [15], Rimon and Koditschek [26], and Connolly and Grupen [6] soon after. The idea is, given the target point, to construct a scalar function on the domain, such that a path to the target point from any other source point may be obtained by following the negative gradient of the function. While elegant, Koren and Borenstein [17] have identified a number of significant pitfalls that these methods may encounter, the most important being the presence of so-called "trap" situations – the presence of local minima in the potential function. To avoid this, the scalar function must have a global minimum (typically zero-valued) at the target, and be void of local minima elsewhere in the domain. The presence of "spurious" local minima could be fatal, since the gradient vanishes and the agent becomes "stuck" there. Other critical points, such as saddles, are undesirable but not fatal, since a negative gradient can still be detected by "probing" around the point.

Designing and computing potential functions for planar domains containing obstacles has been a topic of intense activity for decades. Perhaps the most elegant type of potential function is the harmonic function [15, 6], which has very useful mathematical properties, most notably the guaranteed absence of spurious local minima. Alas, the main problems preventing widespread use of these types of potential functions are the high complexity of computing the function, essentially the solution of a very large system of linear equations, and the fact that very high precision numerical methods are required, as the functions are almost constant, especially in regions distant from the target. This paper addresses the first of these issues. We describe a family of new functions, which, while quite distinct from the harmonic potential function, generate *exactly* the same gradient-descent paths. However, they do this at a tiny fraction of the computational cost.

R. Chen is with Max Planck Institute for Informatics, Saarbrucken, Germany (e-mail: renjie.c@gmail.com)
C. Gotsman is with Jacobs Technion-Cornell Institute, Cornell Tech, New York, USA (e-mail: gotsman@cs.technion.ac.il)
K. Hormann is with Università della Svizzera italiana, Lugano, Switzerland (e-mail: kai.hormann@usi.ch)

In practical path planning scenarios, the planar domain is described by a set of polygons representing the domain boundary and the obstacles within, which can be quite complicated. A good potential function should be "shape-aware", in the sense that it should produce paths which naturally circumvent the obstacles. The agent is armed with an automatic algorithm relying on auxiliary data structures which, given its current position in the domain, can efficiently compute the direction in which it should proceed towards the target. As we shall see later, there is a tradeoff between space and time complexity in achieving this goal.

Although the classical term is "potential function", in this paper we use the more generic term "distance function" for the guiding scalar function. We believe this is more appropriate, as in a sense, the function measures a scalar distance value between the source and the target, which the path-planner tries to decrease as it advances towards the target. Although not identical to the classical shortest-path distance (also known as "geodesic" distance), this distance also takes into account the geometry of the domain and the obstacles.

The rest of this paper is organized as follows. We start with a mathematical analysis of a number of distance functions on continuous planar domains: In Section II we consider distance functions used for path-planning which are common in the literature, based on various forms of the Laplace operator, including the classical harmonic potential function. In Section III we introduce our new family of divergence distance functions, and show (in the Appendix) that they are all equivalent to the Green's function and the Poincaré metric on the disk. In Section IV we direct our attention to the more practical case of a discretized planar domain and provide explicit algebraic expressions and computation methods for the distance functions. There we show how divergence distance functions may be computed much faster than any of the traditional distances. In Section V we provide more experimental results and insights. We conclude in Section VI with a summary and open questions.

## II. LAPLACIAN-BASED DISTANCE FUNCTIONS

**The Green's function**

Classical potential functions are based on harmonic functions, which satisfy the second-order linear differential Laplace equation

$$\nabla^2 f = 0 \qquad (1)$$

where $\nabla^2 = \frac{\partial^2}{\partial x^2} + \frac{\partial^2}{\partial y^2}$ is the Laplace operator, also called the *Laplacian*. These are particularly attractive since a harmonic function satisfies a "minimum/maximum principle" - it obtains its minimum and maximum on the domain boundary, implying that the domain interior contains no local extrema. Beyond this, harmonic functions have many other "nice" properties and have been studied extensively for decades. Rather than providing a detailed exposition here, we refer the interested reader to the book by Axler et al. [1] and the related text by Garnett and Marshall [11], which contain a wealth of information, including all the classical results we use here. Denote by $\Omega$ the open domain, by $\partial\Omega$ its boundary, by $q$ the source point, and by $p$ the target point. Note that the mathematical translation of "obstacles" in the domain, is to "holes" in $\Omega$. If there are no obstacles, $\Omega$ is simply connected and has a single exterior boundary loop. If there are obstacles, $\Omega$ is multiply connected having a single exterior boundary loop and multiple interior boundary loops. For convenience, we identify the plane $R^2$ with the complex field $\mathbb{C}$, and much of our notation and formula will use complex number algebra. For example, a point $(x, y) \in R^2$ is identified with the point $z = x + iy \in \mathbb{C}$, its conjugate is $\bar{z} = x - iy$, its absolute value is $|z| = \sqrt{z\bar{z}} = \sqrt{x^2 + y^2}$, its *Wirtinger derivatives* are $\frac{\partial}{\partial z} = \frac{1}{2}\left(\frac{\partial}{\partial x} - i\frac{\partial}{\partial y}\right)$, $\frac{\partial}{\partial \bar{z}} = \frac{1}{2}\left(\frac{\partial}{\partial x} + i\frac{\partial}{\partial y}\right)$, the vector gradient operator is $\nabla = 2\frac{\overline{\partial}}{\partial z}$ and the scalar Laplacian is $\nabla^2 = 4\frac{\partial^2}{\partial z \partial \bar{z}}$.

The harmonic potential function is derived from the classical *Green's function*, which plays an important role as a *kernel* function in solving the Poisson equation. A Green's function $D$ with Dirichlet boundary conditions is a symmetric scalar bivariate function on the domain:

$$\nabla^2 D(w, z) = \delta(w - z) \qquad \forall\, w, z \in \Omega \qquad (2)$$
$$s.t. \quad D(w, z) = 0 \qquad \forall\, w \in \partial\Omega \text{ or } z \in \partial\Omega$$

where the Laplacian may be applied to either of the two variables.

In the plane with no boundary, it is well known that the Green's function (or *fundamental solution*) is the logarithmic potential:

$$D(w, z) = -\frac{1}{2\pi}\log|w - z|$$

When $\Omega$ has a boundary, it is also well known that for any fixed $w$, $D$ may be written as:

$$D(w, z) = h_w(z) - \frac{1}{2\pi}\log|w - z| \qquad (3)$$

where $h_w$ is a harmonic function of $z$ on $\Omega$. We say that the resulting $D$ has a *pole* at $w$. In practice, $h_w$ is obtained by solving the following Laplace equation with Dirichlet boundary conditions:

$$\nabla^2 h_w(z) = 0 \quad \forall\, z \in \Omega \qquad (4)$$
$$s.t. \quad h_w(z) = \frac{1}{2\pi}\log|w - z| \qquad \forall\, z \in \partial\Omega$$

Given a target $p$, the potential function, whose negative gradient is followed, is actually the negative of $D(p, z)$, which is harmonic on all of $\Omega$ except $p$, and goes from 0 on the boundary to $-\infty$ at $p$ Note the very important fact that $D$ must be computed anew if $p$ is changed. See an example of a Dirichlet Green's function and the gradient-descent path it generates in Fig. 1 (top left). These paths tend to stay away from boundaries. Note that a Dirichlet Green's function will typically be very flat (i.e. almost constant) away from the target point, which, as we will see later, may cause numerical issues when computed.

An interesting variant of the potential function is the lesser-known Green's function with Neumann boundary conditions - $N$ (also called the *Neumann Green's function* or just *Neumann's function* of the domain). Here the situation is a little more complicated [10]: the Laplace equation becomes the more general Poisson equation and the vanishing boundary conditions are replaced by vanishing *normal derivatives* on the boundary. By the normal derivative of a function, we mean the

component of its gradient vector in the outward direction normal to the boundary:

$$\nabla^2 N(w,z) = \delta(z-w) - \frac{1}{A} \quad (5)$$
$$s.t. \quad \frac{\partial N}{\partial n}(w,z) = 0 \quad \forall w \in \partial\Omega \text{ or } z \in \partial\Omega$$

where $A$ is the area of $\Omega$. It is obvious from (5) that for all $w \neq z \in \Omega$, $N(w,z)$ is superharmonic (i.e. has a positive Laplacian) in $z$, and is subharmonic for $z = w$. Since superharmonic functions also have a minimum principle [1], $N(w,z)$ has no local minima in $\Omega$. It is less obvious, but rather easy to prove, that

$$\forall w \in \Omega, \quad \iint_\Omega N(w,z) da_z = 0$$

where $da_z$ is the area element by $z$. As opposed to the Dirichlet Green's function $D$, the Neumann Green's function $N$ will not be symmetric and will have a healthy spread of positive and negative values over the domain. Another important distinction between the paths generated by the Neumann Green's and those generated by the Dirichlet Green's function is that the Neumann Green's path tends to be attracted to domain boundaries. An example of a Neumann Green's function and the gradient-descent path it generates may be seen in Fig. 1 (bottom left).

We conclude by noting that it is possible to build *hybrid* Green's functions by mixing Dirichlet and Neumann boundary conditions to form so-called *Robin's* boundary conditions. The interested reader is referred to Garrido et al. [12] for a qualitative comparison of the gradient-descent paths generated by these types of Green functions and the pure Dirichlet and Neumann cases.

**The heat kernel**
Just as the Green's function, which is a kernel function for the Poisson equation, may be used for path planning, other kernels for more complicated linear differential equations may also be used. A useful kernel is the so-called *heat kernel* [3], which is the generic solution to the heat equation, having two spatial parameters $z$ and $w$ and a temporal parameter $t$:

$$\frac{\partial H}{\partial t}(w,z,t) = \nabla^2 H(w,z,t) \quad (6)$$
$$\lim_{t \to 0} H(w,z,t) = \delta(w-z)$$

In the plane with no boundary, the heat kernel (or fundamental solution) is the Gaussian:

$$H(w,z,t) = \frac{1}{2\pi t} \exp\left(-\frac{|w-z|^2}{4t}\right)$$

When $\Omega$ has a boundary, the heat kernel of $\Omega$, as with the Green's functions, may be defined with either Dirichlet or Neumann boundary conditions similarly to (2) or (5).

The heat kernel turns out to have properties which make it useful as a distance function. As shown by Varadhan [28], for a given target point $p$ and very small $t$, the gradients of $H(z,p,t)$ parallel the gradients of the domain geodesics to $p$. Crane et al. [7] take advantage of this to efficiently compute geodesic distances within a domain. In practice, $H(z,p,t)$ may be computed for a given $t$ and target $p$ by solving the following differential equation, in which the derivative by $t$ has been approximated by a single backward Euler step:

$$H(z) - t\nabla^2 H(z) = \delta(z-p) \quad (7)$$
$$H(z) = 0 \quad \forall z \in \partial\Omega \quad \text{(Dirichlet)}$$
$$\frac{\partial H}{\partial n}(z) = 0 \quad \forall z \in \partial\Omega \quad \text{(Neumann)}$$

Conveniently, the solution to this particular differential equation also has a maximum principle [25], hence the distance function, which is actually the negative of $H$, is void of local minima within the domain. We denote by $HD$ the distance function based on Dirichlet boundary conditions, and by $HN$ that based on Neumann boundary conditions. See an example of $HD$ and $HN$ and the paths they generate for $t = 10^{-3}$ in the second column of Fig. 1. Note that both decay very rapidly away from the target, which, as with the Dirichlet Green's function, may lead to numerical issues. The main qualitative difference between the two, as with the Greens' functions, is that $HD$ tends to be repelled from the boundaries and $HN$ tends to be attracted to boundaries. However, as noted by Crane et al. [7], in regions distant from any boundary their behavior is very similar. The observant reader will notice that the Green's functions may also be obtained as a special case of the solution to (7) as $t \to \infty$.

**Spectral distances**
A number of other shape-aware distance functions have been proposed in the literature based on the eigenstructure of the Laplacian on the domain. The generic form of such a distance function is:

$$S(w,z) = \sum_{k=1}^{\infty} f(\lambda_k)\big(\phi_k(w) - \phi_k(z)\big)^2 \quad (8)$$

where $(\phi_k, \lambda_k)$ is the $k$-th eigenfunction and eigenvalue of $\nabla^2$. The most well-known such distance function is the so-called *diffusion* distance [5], which, similarly to the heat kernel-based distance, has a temporal parameter $t$. Here $f(x) = \exp(-2tx)$. Other distances are the *resistance* distance $f(x) = 1/x$ and the *biharmonic* distance $f(x) = \frac{1}{x^2}$ defined by Lipman et al. [19]. The diffusion distance is equivalent to the $L_2$ distance between two heat kernels with parameter $t$ centered at $z$ and $w$, respectively. The resistance distance may not be well-defined using (8), since the infinite sum may not converge, so the correct way to define it is as an *extremal length*, which is known to be a conformal invariant [9]. As we shall see later, the resistance distance has a very simple equivalent definition in the discrete case [16], which has been used extensively in network analysis. For an extensive discussion and comparison between these spectral distances, the interested reader is referred to Patanè and Spagnuolo [22].

The resistance and biharmonic distances may also be defined in terms of the kernel function of their respective linear differential operators, the *Laplacian* $\nabla^2$ or the *bi-Laplacian* $\nabla^4$, namely, if ♦ is the operator and $K$ is the solution to:

$$\blacklozenge K(w,z) = \delta(z-w) \quad (9)$$
$$\frac{\partial K}{\partial n}(w,z) = 0 \quad \forall w \in \partial\Omega \text{ or } z \in \partial\Omega \quad \text{(Neumann)}$$

then the distance is:
$$d(w,z) = K(z,z) + K(w,w) - 2K(w,z) \quad (10)$$

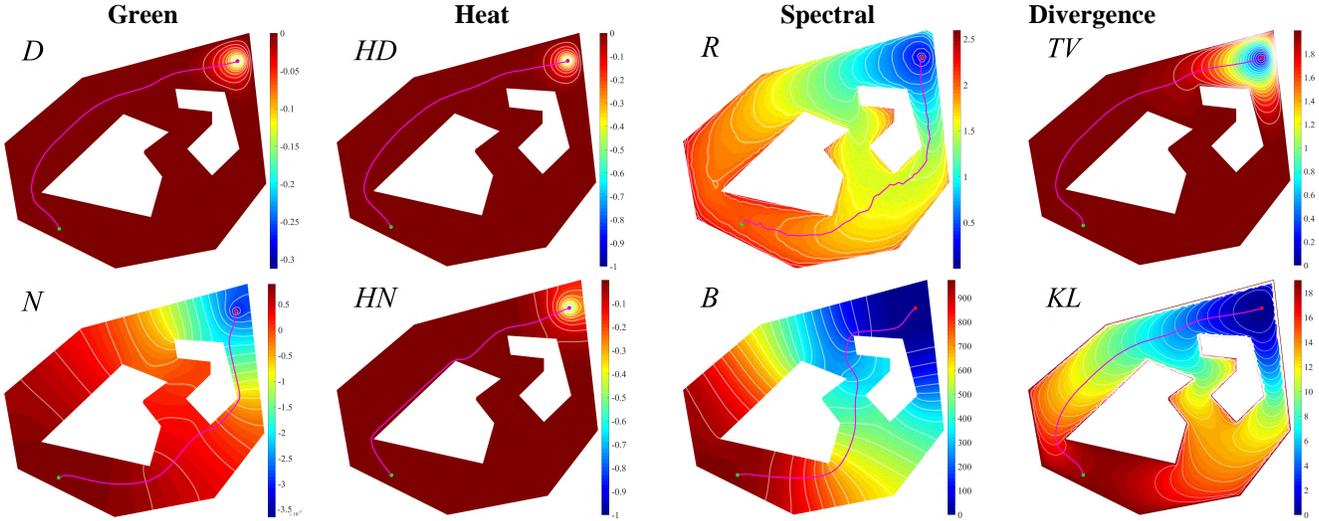

Fig. 1. Paths from green source point to red target point generated by following the negative gradient of various distance functions in the Convex Holes domain. **(left)** Green's function with (top) $D$ = Dirichlet and (bottom) $N$ = Neumann boundary conditions. **(second from left)** Heat kernel using $t=10^{-3}$ with (top) $HD$ = Dirichlet and (bottom) $HN$ = Neumann boundary conditions. **(second from right)** Spectral distances. (top) Resistance distance $R$. (bottom) Biharmonic distance $B$ (with Neumann boundary conditions). Domain is color-coded with function value. White curves are function contours (level sets). Note how these are parallel to the boundary in the Dirichlet case and normal to the boundary in the Neumann case. **(right)** Our divergence distance functions: (top) $TV$, (bottom) $KL$. Note that both paths are identical to the path generated by $D$.

Note that for the resistance distance $K$ is no other than the Green's function. We will denote the resistance distance by $R$ and the biharmonic distance by $B$. Due to the infinite values $K(z,z)$ and $K(w,w)$ of the Green's function, defining $R$ through (10) is quite unstable, resulting in noisy distances, as can be seen in the third column of Fig. 1. In contrast, defining $B$ through (10) works well in most cases, although this distance has two distinct disadvantages: (1) it does not come with a guarantee of no spurious local minima, and (2) it is quite expensive to compute for a given target $w=p$, since $K(z,z)$ is required for any $z$. As demonstrated by Lipman et al [19], the spectral definition (8) implies an efficient computation of a spectral distance by discarding the eigenvectors with small eigenvalues, but this is still quite expensive. The third column of Fig. 1 shows the gradient-descent path generated by the biharmonic distance.

### III. Divergence Distances

In this section we introduce a new family of distance functions which, although different from all the distance functions mentioned in the previous section, generate gradient-descent paths *identical* to the Dirichlet Green's function $D$. When this happens, we say that the distance functions are *equivalent* to $D$. It happens because their gradient vector field is identical in *direction*, but not in magnitude, to the gradient vector field of $D$. Moreover, as we will see, the new distance functions are much easier to compute. Our point of departure is Green's theorem [1], which implies that the unique harmonic function $h(z)$ on $\Omega$ having given boundary conditions $b(\partial\Omega)$ may be computed as the following boundary integral:

$$h(z) = \oint_{\partial\Omega} \frac{\partial D}{\partial n}(s,z) b(s) ds \qquad (11)$$

where, as in the previous section, $D$ is the Dirichlet Green's function and $\frac{\partial D}{\partial n}$ is its normal derivate at the boundary point $s$. The kernel function $P(s,z) = \frac{\partial D}{\partial n}(s,z)$ relating boundary points $s$ to interior points $z$ is called the *Poisson kernel* [1] of $\Omega$. Strictly speaking, the term Poisson kernel is usually used when the domain is the unit disk, but here we will be more liberal with the terminology. $P(s,z)$ may also be obtained directly as the solution to:

$$\nabla^2 P(s,z) = 0 \qquad \forall z \in \Omega \qquad (12)$$
$$P(s,t) = \delta(s-t) \qquad \forall s,t \in \partial\Omega$$

from which it is easy to see that

$$\oint_{\partial\Omega} P(s,z) ds = 1 \qquad \forall z \in \Omega \qquad (13)$$

meaning that for all $z \in \Omega$, $P(s,z)$ may be thought of as a probability density on $\partial\Omega$. For a subset $E$ of the boundary, the quantity

$$\omega(E,\Omega) = \oint_E P(s,z) ds \qquad E \subset \partial\Omega$$

is called the *harmonic measure* of $E$ [9], and can be interpreted as the probability that a random walk in $\Omega$ starting at $z$ will exit $\Omega$ for the first time through $E$.

Conversely, if we know the Poisson kernel of $\Omega$, the Dirichlet Green's function may be obtained from (3) and (11) as the boundary integral:

$$D(w,z) = -\frac{1}{2\pi}\log|z-w|$$
$$+ \frac{1}{2\pi}\oint_{\partial\Omega} P(s,z) \log|z-s| ds \qquad (14)$$

To illustrate, we note that in the special case where $\Omega$ is the unit disk, we have closed formulae for the Green's functions and Poisson kernel:

$$D(w,z) = -\frac{1}{2\pi}\log\frac{|w-z|}{|1-\bar{z}w|}$$

$$N(w,z) = -\frac{1}{2\pi}\log(|w-z||1-\bar{z}w|) \quad (15)$$

$$P(s,z) = \frac{1}{2\pi}\frac{1-|z|^2}{|s-z|^2}$$

Unfortunately, for any other non-trivial domain, there are no closed formula for these functions, and they are typically computed numerically, as we will elaborate on in the next section.

We now define a new family of distance functions using boundary integrals. These are based on the concept of $f$-divergence introduced by Csiszár [8], which is used in statistics and information theory as a way to measure the difference between two probability measures. Assume that $f:\mathbb{R}^+ \to \mathbb{R}$ is a continuous convex function such that $f(1) = 0$. The *divergence distance* between $w$ and $z$ is the $f$-divergence of the two harmonic measures $P(s,z)$ and $P(s,w)$:

$$DV(z,w) = \oint_{\partial\Omega} P(s,z) f\left(\frac{P(s,w)}{P(s,z)}\right) ds$$

The most important property of $DV$ motivates its use as a distance function for our purposes:

$$DV(z,w) \geq 0 \quad \text{and} \quad DV(z,w) = 0 \text{ iff } z = w$$

In the Appendix we provide an extensive mathematical analysis of divergence distances, and the most important result there is the Equivalence Theorem (Theorem A6), which proves that all distance functions based on *f*-divergences are equivalent to the Dirichlet Green's function, i.e. generate exactly the same gradient-descent paths. The advantage, however, of working with divergence distances is that they may be computed much more efficiently. Computation of $DV$ involves a pre-processing stage where the Poisson kernels $P(t,\cdot)$ are computed for all $t$. Then, given arbitrary $z$ and $w$, computing $DV(z,w)$ becomes a simple integral.

To be concrete, in the sequel we focus on two specific members of the divergence distance family which we believe are sufficiently different and representative: The first, called the *Total Variation* divergence, corresponds to $f(x) = |1-x|$, and is simply the $L_1$ distance between the values of the Poisson kernel on $z$ and $w$:

$$TV(w,z) = \oint_{\partial\Omega} |P(s,z) - P(s,w)| ds \quad (16)$$

The range of $TV$ is $[0,2]$. The second distance function is the *Kullback-Leibler* divergence [18] between the Poisson kernels:

$$KL(w,z) = \oint_{\partial\Omega} P(s,z) \log\frac{P(s,z)}{P(s,w)} ds \quad (17)$$

This divergence (sometimes called the *discrimination information* or *relative entropy*) corresponds to $f(x) = -\log x$, and its range is $[0,\infty)$.

In contrast to $D$, which is a harmonic function everywhere except at the target point, all divergence distances are *subharmonic*, i.e. have non-negative Laplacians. This is a direct consequence of the convexity of $f$:

$$\nabla_w^2 DV(w,z) = \frac{\partial^2}{\partial w \partial \bar{w}} \oint_{\partial\Omega} P(s,z) f\left(\frac{P(s,w)}{P(s,z)}\right) ds$$

$$= \oint_{\partial\Omega} \frac{\partial^2}{\partial w \partial \bar{w}}\left(P(s,z) f\left(\frac{P(s,w)}{P(s,z)}\right)\right) ds$$

$$= \oint_{\partial\Omega} \left(\frac{1}{P(s,z)} f''\left(\frac{P(s,w)}{P(s,z)}\right)\left|\frac{\partial P(s,w)}{\partial w}\right|^2 \right.$$

$$\left. + f'\left(\frac{P(s,w)}{P(s,z)}\right)\frac{\partial P^2(s,w)}{\partial w \partial \bar{w}}\right) ds$$

$$= \oint_{\partial\Omega} \left(\frac{1}{P(s,z)} f''\left(\frac{P(s,w)}{P(s,z)}\right)\left|\frac{\partial P(s,w)}{\partial w}\right|^2 + 0\right) ds \geq 0$$

The last equality is because $P(s,w)$ is a real harmonic function of $w$. In general, a subharmonic distance function could be bad news, as a subharmonic function possesses only a maximum principle, which prevents the occurrence of spurious local *maxima*, but not of spurious local *minima*. However, as the Equivalence Theorem shows, $DV$ is *equivalent* to $D$, namely, has no spurious extrema and has gradient vector fields identical in *direction* to the gradient vector field of $D$. Consequently, they generate gradient-descent paths which are *identical* to those generated by $D$, as demonstrated in Fig. 1. The proof of the Equivalence Theorem is based on the fact that all three distance functions are *conformal invariants* of the domain.

Furthermore, Section A2 of the Appendix shows that in the special case of the unit disk, the paths generated by $D$ and $DV$ are hyperbolic circles, which are symmetric in the source and the target, namely, swapping the roles of these two points results in the same path, which is a desirable feature, and carries over to general domains. The same cannot be said for the other distances not equivalent to $D$. Some of these paths for the unit disk are shown in Fig. A1 in the Appendix.

IV. DISCRETE DISTANCE FUNCTIONS

Sections II and III described a number of scalar functions, all based on the Laplacian, which could be used as domain-aware distance functions to generate gradient-descent paths to a given target point in the domain. While $D, N, HD, HN, TV$ and $KL$ have been shown to have the necessary properties, the proof relies heavily on the fact that the domain is continuous and it is not obvious that these properties carry over to practical implementations where the domain is discretized. In this section, we explore the discrete setting. As in the continuous setting, the key player is the Laplacian and the related notion of harmonicity, which fortunately is well defined also in the discrete setting.

Assume a planar domain $\Omega$ discretized into finite elements by a triangulation $T$ having $n$ vertices in its vertex set $V$, of which $k$ are in the set $B$ of boundary vertices and $m$ are in the set $I$ of interior vertices. The discrete Laplacian of $\Omega$ is a linear operator on the triangulation vertices – a sparse symmetric $n \times n$ matrix $L^c$ whose non-zero entries corresponding to triangulation edges are given by the cotangent formula [23], and each diagonal entry is the negative of the sum of the corresponding row entries:

$$L_{ij}^c = \frac{1}{2}(\cot \alpha_{ij} + \cot \beta_{ij}) \quad i \neq j$$

$$L_{ii}^c = -\sum_j L_{ij}^c$$

where $\alpha_{ij}$ and $\beta_{ij}$ are the angles opposite the edge between vertices $i$ and $j$.

This formula is the classical *conformal* Laplacian. It is well known that if $T$ is a *Delaunay triangulation* [2] of the domain, the entries of $L$ corresponding to interior edges are guaranteed to be positive, which is a critical property of a Laplacian.

For the matrix to best approximate the continuous Laplacian in the plane, a more accurate version is the non-symmetric matrix:

$$L_{ij} = \frac{1}{2a_i}(\cot \alpha_{ij} + \cot \beta_{ij})$$
$$L_{ii} = -\sum_j L_{ij}$$

where $a_i$ is the area of the dual *Voronoi cell* [2] of the $i$-th vertex. In practice, $a_i$ is approximated as one third of the areas of the triangles incident on the $i$-th vertex. In a more compact form, this can be written as:

$$L = A^{-1}L^c, \quad A = diag(a)$$

where $a = (a_1, ..., a_n)$ is a vector of the areas. The literature also contains a "normalized" non-symmetric version of the Laplacian:

$$L^n = Z^{-1}L^c, \quad Z = diag(L^c)$$

which is common in probabilistic scenarios, e.g. random walk theory. Obviously $L^n$ has a unit diagonal.

If, without loss of generality, we assume the following block structure for $L$:

$$L = \begin{pmatrix} L_{II} & L_{IB} \\ L_{BI} & L_{BB} \end{pmatrix} \quad (18)$$

then the notion of harmonicity of a real function $f$ defined on the vertices of the triangulation (represented as a vector) is:

$$L_{II}f_I + L_{IB}f_B = 0$$

where $f_I$ is the subvector of $f$ corresponding to the interior vertices and $f_B$ is the subvector of $f$ corresponding to the boundary vertices.

### The Dirichlet Green's function

The Green's function of $T$ with Dirichlet boundary conditions is the function (matrix) $D: V \times V \to R$, such that:

$$(L_{II} \ L_{IB})\begin{pmatrix} D_{II} \\ D_{BI} \end{pmatrix} = I \text{ harmonic at interior vertices except "poles"} \quad (19)$$
$$D_{BI} = D_{IB}^T = \mathbf{0} \quad \text{vanishing boundary conditions}$$

The main difference between this definition and that of the continuous case (1) or (2) is that the matrix $D$ is not neccesarily symmetric and its values are all finite, also on the diagonal (since the continuous singular Dirac delta function is replaced by the discrete finite Kronecker delta). It is easy to see that the unique solution to (19) is the $m \times m$ matrix:

$$D_{II} = L_{II}^{-1} \quad (20)$$

so, given an interior vertex $p \in I$, it is possible to compute $D_{Ip}$ — the Green's function of $T$ with pole at $p$ — or alternatively, the $p$-th column (or row) of $D$ — by solving the following linear system:

$$L_{II}D_{Ip} = \begin{pmatrix} L_{JJ} & L_{Jp} \\ L_{pJ} & L_{pp} \end{pmatrix}\begin{pmatrix} D_{Jp} \\ D_{pp} \end{pmatrix} = \begin{pmatrix} \mathbf{0} \\ 1 \end{pmatrix} \quad (21)$$

where $J$ is the set of interior vertices excluding $p$, and $\mathbf{0}$ is a column vector of $M - 1$ zeros, whose solution is:

$$D_{pp} = \frac{1}{L_{pp} - L_{pJ}L_{JJ}^{-1}L_{Jp}}, \quad D_{Jp} = -L_{JJ}^{-1}L_{Jp}D_{pp} \quad (22)$$

It is interesting to note that $D_{Ip}$ is equivalent to the *harmonic potential function* $Q_{Jp}$ obtained by solving the Laplace equation on $T$ with boundary conditions $Q_{pp} = 1$ and $Q_{Bp} = \mathbf{0}$:

$$(L_{JJ} \ L_{JB} \ L_{Jp})\begin{pmatrix} Q_{Jp} \\ \mathbf{0} \\ 1 \end{pmatrix} = \mathbf{0}$$

whose solution is:

$$Q_{Jp} = -L_{JJ}^{-1}L_{Jp} \quad (23)$$

which is identical to $D_{Jp}$ in (22) up to the constant factor $D_{pp}$. Both are discrete harmonic on $J$. In practice, we obtain $D_{Ip}$ by directly solving (21). The advantage of this is that Cholesky pre-factorization of $L_{II}$ enables to efficiently solve (21) for different $p$'s without inverting $L_{II}$ by back-substitution [24]. This is because changing the identity of $p$ changes only the right hand side of the equation.

It is also interesting to note that using either of $L, L^c$ or $L^n$ when computing $D$ or $Q$ merely changes the individual columns of these matrices by constant factors, giving essentially the same result for all target vertices. It is computationally advantageous to use the symmetric $L^c$ matrix, as this allows using efficient sparse Cholesky factorization.

### The Neumann Green's function

With Neumann boundary conditions, the requirement of vanishing boundary values is replaced by the requirement of vanishing "normal derivative" values at the boundary vertices. In practice, in the discrete setting, this is approximated by the extension of the harmonic requirement also to the boundary vertices, effectively eliminating any boundary conditions. Thus, in principle, we should have

$$N = L^{-1}$$

but since $L$ is singular, the closest alternative is:

$$N = L^+ = (L - E)^{-1} + E^T, \quad E = \frac{a}{|a|}e^T \quad (24)$$

where $L^+$ is the pseudo-inverse of $L$, $a$ is a column vector of the vertex areas and $e$ is a constant column vector of $\frac{1}{\sqrt{n}}$, which is equivalent to:

$$LN = I - EE^T = I - \frac{aa^T}{|a|^2}$$

Unfortunately, this system is singular because of $L$, but recalling that the rows of $L$ sum to 0, it is equivalent to the non-singular system

$$\begin{pmatrix} L_{JV} \\ \mathbf{1} \end{pmatrix}N = \begin{pmatrix} I_{JV} - \frac{a_J a^T}{|a|^2} \\ 0 \end{pmatrix} \quad (25)$$

where $J$ is the set of all vertices excluding an arbitrary vertex $r$. For a given target vertex $p \neq r$, we immediately see that $N_{Vp}$ — the $p$-th column of $N$ – is superharmonic at every vertex but $p$, where it is subharmonic. $N_{Vp}$ is obtained from (25) by solving a slightly simpler but symmetric system (for $M$)

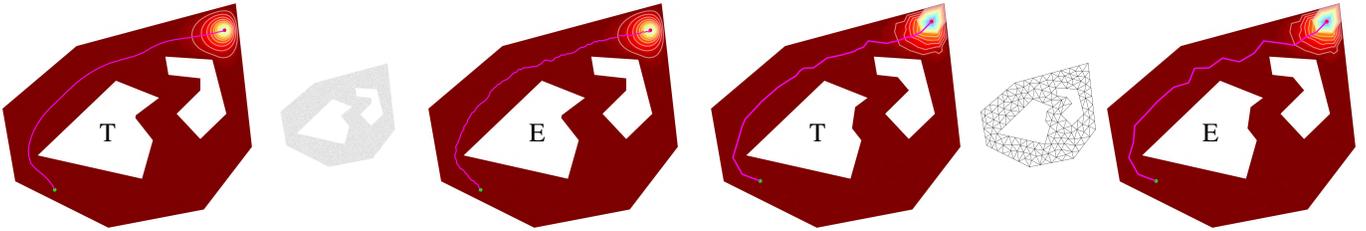

Fig. 2. Triangle-based (T) vs. edge-based paths (E) in a triangle mesh of the Convex Holes domain at high and low resolutions: (Left) $n = 5,815$ and (Right) $n = 175$. Note how the triangle-based paths are much smoother than the edge-based paths.

$$L_{JJ}^C M = a_p \left( \delta_p - \frac{a_J^2}{|a|^2} \right), \qquad N_{Vp} = (I_{VJ} - e e_J^T) M \quad (26)$$

where $a_J^2$ is the vector of squared entries of $a_J$. The advantage of using (26) is that only the right hand side depends on $p$, therefore similarly to (21), the system may be solved efficiently for different $p$'s without inverting $L_{JJ}^C$ by Cholesky pre-factorization of $L_{JJ}^C$, and back-substitution for different right hand sides. The only caveat is that $r$ can never be the target vertex.

It is worthy to note that despite the implicit Neumann boundary conditions, the normal derivative of $N$, as computed in (24) or (25), does not vanish on the boundary. For example, when the domain is the unit disk and $p = 0$, (15) implies that $D$ and $N$ are identical and their normal derivative on the boundary is $\frac{1}{\pi}$. When $t$ moves away from the origin, the normal derivative of $N$ on the boundary diminishes. The left column of Fig. 3 shows gradient-descent paths generated by $D$ and $N$ for a low-resolution triangulation of the domain of Fig. 1.

**Gradient-descent path generation**
In the discrete setting, the source point $q$ and target point $p$ are vertices of the triangulation and the distance function is a scalar per triangulation vertex. The simplest way to generate a gradient-descent path from $q$ to $p$ is to move from a vertex to the neighbor of that vertex which decreases the distance value the most (assuming this neighbor exists because it is not a local minimum). The resulting path is then along edges from $q$ to $p$. However, a smoother path may be generated by cutting through triangles, at the price of a more complex computation. This is achieved by realizing that the three values of the distance function on a triangle define a unique gradient vector for that triangle. Starting at $q$, the steepest gradient at an adjacent triangle may be followed into that triangle, exiting at one of its edges into a new triangle. (For some vertices, especially in a low-resolution mesh, such an adjacent triangle may not exist. In that case, we simply pick the adjacent edge having the largest gradient of the distance, follow it to an adjacent vertex and start the process there.) The gradient of the new triangle may then be followed from that point, and the process repeats. If the path intersects an edge incident on the target, the path terminates along that edge at the target. As with the simple edge-based path tracing procedure, should any of the vertices in the vicinity of the path be a local minimum, this process may get stuck. Assuming that this does not happen, the fact that $p$ is a global minimum guarantees that this path will eventually reach $p$. Fig. 2 compares between the two types of paths. All the paths in the figures in the sequel relating to the discrete case were generated using the second method.

**The heat kernel distance**
Computing the heat kernel distance in the discrete setting involves solving a linear system, the equivalent to (7):
$$H - tLH = I \quad (27)$$
The Dirichlet boundary conditions
$$H_{IB} = H_{BI}^T = \mathbf{0}, \quad H_{BB} = \mathbf{0}$$
lead to:
$$HD_{II} = (I - tL_{II})^{-1} \quad (28)$$
and the Neumann boundary conditions lead to:
$$HN = (I - tL)^{-1} \quad (29)$$
As pointed out by Crane et al. [7] (and is the case with Green's functions), the resulting linear system is equivalent (up to a constant factor) to a linear system that can be pre-factored and back-substituted on demand for different $p$'s. The second column of Fig. 3 shows the resulting gradient-descent paths on a triangulation as compared to those generated by the discrete Green's functions. As mentioned above, the Green's function are obtained when $t \to \infty$ (and in this case $H$ is no longer full rank, thus the pseudo-inverse should be used).

**The spectral distances**
Resistance distance is defined between two vertices in a graph as the effective resistance between these two vertices if each edge is modeled as a unit resistor [16]. This is sometimes called the *commute time*, as it is equivalent to the expected time to perform a random walk from one vertex to the other and back (averaged over all possible paths). The analog in the continuous case is called *extremal length* and is known to be a conformal invariant [9].

In the discrete setting, if we view the triangulation as a planar graph, the resistance distance between vertices $p$ and $q$ has the following simple expression [19]:
$$R_{pq} = G_{pp} + G_{qq} - 2G_{pq}, \qquad G = (L^c)^+ \quad (30)$$
In contrast to both Green's functions, the resistance distance cannot be solved efficiently for a given $p$, since the diagonal value $G_{qq}$ is required for the different $q$'s, which in turns implies that the entire matrix $G$ is required, as it is difficult to compute the diagonal without computing the entire matrix.

Similarly, the discrete biharmonic distance between $p$ and $q$ has the following analogous expression using the pseudo-inverse of the squared Laplacian [19]:
$$B_{pq} = S_{pp} + S_{qq} - 2S_{pq}, \qquad S = (L^c A^{-1} L^c)^+ \quad (31)$$

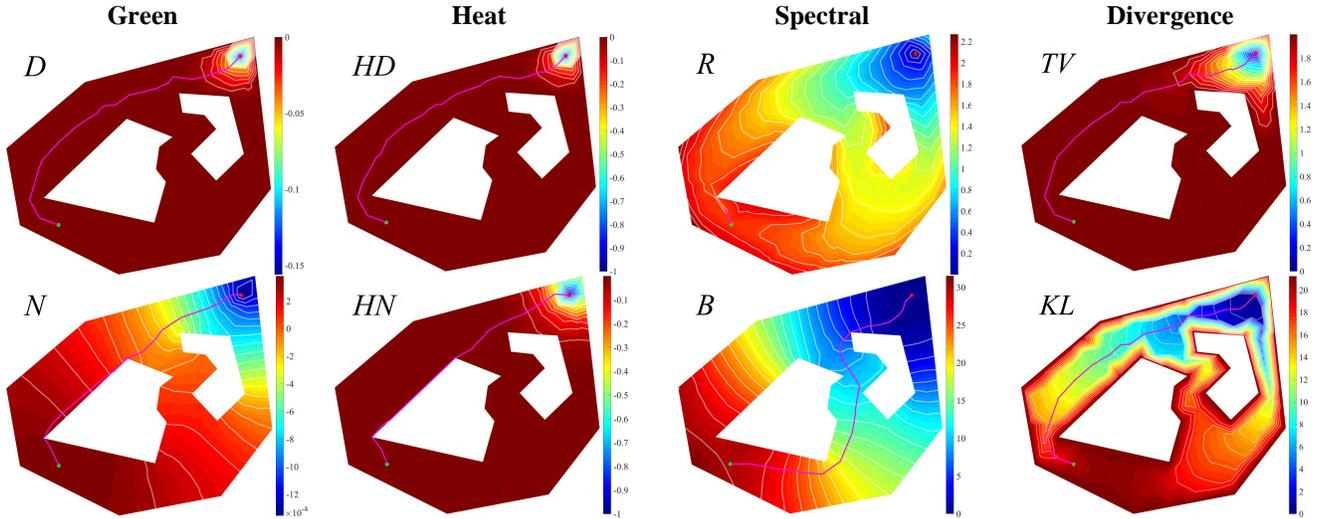

Fig. 3. The discrete setting, where the Convex Holes domain is represented by a low resolution Delaunay triangulation with $n = 175$. The paths from green source point to red target point were generated by following through triangles the negative (piecewise-constant) gradient of the distance function from the target. **(left)** Green's function. (top) $D$ = Dirichlet boundary conditions. (bottom) $N$ = Neumann boundary conditions. **(second from left)** Heat kernel using $t = 10^{-3}$. (top) $HD$ = Dirichlet boundary conditions. (bottom) $HN$ = Neumann boundary conditions. Domain is color-coded with function value. Curves are piecewise-linear function contours (level sets). **(second from right)** Spectral distances. (top) $R$ = resistance distance, (bottom) $B$ = biharmonic distance (with Neumann boundary conditions). Note that the path based on resistance distance encounters a local minimum almost immediately, thus gets "stuck". **(right)** Divergence distances: (top) $TV$ and (bottom) $KL$. Domain is color-coded with function value. Curves are function contours (level sets). Note that the paths generated by $TV$ and $KL$ are visually identical to the path generated by $D$.

**The divergence distances**

As described above, the $TV$ and $KL$ distances are defined via the Poisson kernel. In the discrete setting, the Poisson kernel is a $n \times (n - m)$ non-negative matrix $P$ with unit sum rows (one "Poisson kernel" vector for each triangulation vertex). The $i$-th column of $P$ represents the $i$-th boundary vertex, and is a harmonic function on the triangulation with boundary value vector which is the binary indicator vector of that boundary vertex. With the notation of (18), this means

$$(L_{II} \quad L_{IB}) \begin{pmatrix} P_{IB} \\ I \end{pmatrix} = \mathbf{0} \tag{32}$$

where $I$ is the $(n - m) \times (n - m)$ identity matrix. Thus:

$$P = \begin{pmatrix} P_{IB} \\ I \end{pmatrix}, \quad P_{IB} = -L_{II}^{-1} L_{IB} \tag{33}$$

Note that identical results are obtained if $L$, $L^c$ or $L^n$ are used to define $P$. Given the Poisson kernel vectors, the $TV$ and $KL$ distances are $n \times n$ matrices with the following entries:

$$TV = \sum_{b \in B} |P_{pb} - P_{qb}| \tag{34}$$

$$KL_{pq} = \sum_{b \in B} P_{pb} \log \frac{P_{pb}}{P_{qb}} \tag{35}$$

Figure 3 shows the paths generated using the $TV$ and $KL$ distances in the discrete setting. Note that they are identical to the paths generated by $D$ in the same figure.

## V. Experimental Results

We have implemented all the distance functions described in this paper on a variety of domains: convex and non-convex, with and without holes. The domains were discretized at various resolutions using a Delaunay triangulation generated by the *Triangle* software package [27].

**Computational Complexity**

One of the practical problems with the classical Green's functions $D$ and $N$, as with the heat kernel distances $HD$ and $HN$, given a target point $p$, is that they are difficult to compute. In the discrete setting, as elaborated in the previous section, they involve solving a sparse linear system, which may be solved using either iterative methods, e.g. Gauss-Seidel, which are slow but require modest space, or direct methods [24], which are faster but have more storage requirements. Indeed, in the early days, direct methods on large linear systems were prohibitive. With the advent of efficient numerical solvers, sufficient resources and parallel processing hardware, these have become more accessible and a popular method to solve *sparse* linear systems is by prefactoring the positive-definite system matrix using e.g. Cholesky decomposition, and then solving for an arbitrary right hand side using back-substitution, which makes for a relatively efficient $O(n)$ method. However, it also means that for a given target vertex $p$ of the triangulation, the distance function for the *entire* vertex set is solved for, which is obviously unnecessary, as the distance at only the vertices relevant to the path is really required. Using the $TV$ or $KL$ distances avoids this unnecessary computation, as once the Poisson kernel vectors have been computed in a pre-processing stage (independent of the identity of $p$), the boundary sum definition (34), (35) allows to compute the distance between any two vertices relatively efficiently (in $O(\sqrt{n})$ or less, depending on sparsity) on demand. We have implemented prefactorization and backsubstitution required for computing $D$ on the CPU using the Intel Math Kernel Library (MKL) [13] and on the GPU using the NVIDIA cuSOLVER package [21]. Computation of the $KL$ and $TV$ distances on the CPU is straightforward and on the GPU may be done easily in CUDA thanks to its embarrassingly parallel nature.

TABLE I
RUNTIME ON THE CPU. For **TV** and **KL**, only distances to those vertices relevant to the path are computed.

| Domain | #vertices | | D | | TV | | KL | | | | |
|---|---|---|---|---|---|---|---|---|---|---|---|
| | total | boundary | Preproc. (ms) | Online (ms) | Preproc. (ms) | Online (ms) | Preproc. (ms) | Dense online (ms) | Sparse online (ms) | #vertices on path | Sparsity (%) |
| LoRes Man2 | 1,886 | 492 | 0.003 | 0.168 | 0.045 | 0.067 | 0.045 | 0.086 | 0.011 | 219 | 89 |
| HiRes Man2 | 7,117 | 951 | 0.032 | 2.255 | 0.382 | 0.334 | 0.382 | 0.410 | 0.029 | 386 | 91 |
| HiRes Man1 | 6,939 | 903 | 0.021 | 1.929 | 0.371 | 0.292 | 0.371 | 0.338 | 0.036 | 378 | 90 |
| Concave | 8,319 | 499 | 0.040 | 2.873 | 0.309 | 0.121 | 0.309 | 0.143 | 0.016 | 295 | 88 |
| Concave Holes | 6,646 | 567 | 0.019 | 1.489 | 0.268 | 0.143 | 0.268 | 0.152 | 0.016 | 322 | 90 |
| Maze | 18,929 | 6,167 | 0.046 | 2.913 | 5.264 | 8.826 | 5.264 | 8.565 | 0.133 | 1,407 | 99 |
| Convex Holes | 5,815 | 491 | 0.027 | 0.997 | 0.188 | 0.131 | 0.188 | 0.148 | 0.016 | 282 | 84 |
| Disk | 16,002 | 397 | 0.077 | 5.296 | 0.515 | 0.052 | 0.515 | 0.067 | 0.056 | 155 | 0 |

TABLE II
RUNTIME ON THE GPU. From one target vertex to all other vertices.

| Domain | #vertices | | D | | TV | | KL | | | |
|---|---|---|---|---|---|---|---|---|---|---|
| | total | boundary | Preproc. (sec) | Online (ms) | Preproc. (sec) | Online (ms) | Preproc. (sec) | Dense online (ms) | Sparse online (ms) | Sparsity (%) |
| LoRes Man2 | 1,886 | 492 | 0.016 | 2.589 | 1.178 | 0.062 | 1.178 | 0.074 | 0.064 | 89 |
| HiRes Man2 | 7,117 | 951 | 0.129 | 17.792 | 15.298 | 0.089 | 15.298 | 0.052 | 0.063 | 91 |
| HiRes Man1 | 6,939 | 903 | 0.155 | 16.310 | 13.482 | 0.051 | 13.482 | 0.051 | 0.074 | 90 |
| Concave | 8,319 | 499 | 0.248 | 26.196 | 12.328 | 0.052 | 12.328 | 0.052 | 0.057 | 88 |
| Concave Holes | 6,646 | 567 | 0.156 | 18.889 | 10.086 | 0.058 | 10.086 | 0.051 | 0.059 | 90 |
| Maze | 18,929 | 6,167 | 0.094 | 16.902 | 90.525 | 0.081 | 90.525 | 0.063 | 0.059 | 99 |
| Convex Holes | 5,815 | 491 | 0.096 | 15.238 | 6.885 | 0.055 | 6.885 | 0.053 | 0.057 | 84 |
| Disk | 16,002 | 397 | 1.793 | 58.106 | 25.432 | 0.074 | 25.432 | 0.053 | 0.093 | 0 |

When computing on the CPU, the $KL$ and $TV$ distances are computed only for relevant vertices along the generated path, whereas the $D$ distances are inevitably computed for the entire mesh. When computing on the GPU, the *entire* distance field is computed for a given $p$ even for $KL$ and $TV$ since, as in many parallel computing scenarios, it costs almost the same as computing the values only along the path. Furthermore, the sparse nature of the Poisson kernel vectors may be used to accelerate the $KL$ computations. Tables 1 and 2 show CPU and GPU performance timings in computing paths on different domains using the $D$, $KL$ and $TV$ distances. The domains are a variety of shapes and sizes, where the boundary size (in vertices) ranges between 2.5% (the disk) and 33% (the maze) of the total triangle mesh size. Computing $D$ and $KL/TV$ efficiently both require *preprocessing* which is independent of the target vertex $p$, and *online* processing which is dependent on $p$. For $D$ the preprocessing consists of Cholesky decomposition, and the online processing is back-substitution. For $TV$ and $KL$, the preprocessing consists of computation of the Poisson kernel vectors, and the online processing is computing the sums (34) or (35). The tables list also the preprocessing times, but since this is performed only once per input domain, they are of minor importance. The more important figure is the online processing time. When using the CPU, on all domains but the maze (see Fig. 7), we observe a speedup in the online processing time of anywhere between 2x and 100x. On the maze, in which a long path may effectively cover the entire domain, the online computation of $TV$ and $KL$ was 3x *slower* than the online computation of $D$. However, when employing the approximation to $KL$ possible because of the sparsity of the Poisson kernels, a speedup of 22x was obtained for the maze, and anywhere between 15x to 180x on the other domains. The sparsity in this maze was the best (99%) and in the disk was the worst (0%). On the other domains it was in the vicinity of 90%.

When using the GPU, the speedup of the online computation of $TV$ and $KL$ relative to $D$ is quite significant. Interestingly, probably due to the inherent serial nature of both the decomposition and the back-substitution operations, their GPU implementations seem to be much slower than their CPU implementation, further boosting the speedups of the online computation of $TV$ and $KL$ relative to $D$. For $TV$ it is anywhere between 42x and 780x, and for $KL$ anywhere between 35x and 1100x. The sparse approximation for $KL$ did not improve this significantly.

**Precision**

One of the other practical problems with the Dirichlet Green's function $D$ is that it is very flat, almost constant, when far away from the target point, as evident in Figs. 1 and 3. This makes its computation highly sensitive to numerical precision issues and in some cases can even make the result unusable. Wray et. al [25], who use an iterative Gauss-Seidel method to compute $D$, show how to modify the iteration so that essentially $\log D$ is computed with sufficient precision (of course, this new method is far superior to computing $D$ and simply applying the logarithm to the result). It is based on the well-known principle that it is best to compute the logarithm of a product of very small numbers as the sum of the individual logarithms. Our $KL$ distance function, which we have shown to be equivalent to the Dirichlet Green's function, suffers from the same precision problem since the Poisson kernel vectors are computed similarly to $D$. As a remedy, the method of Wray et. al [29] may be applied here as well to obtain the vectors $\log P$. These are easily combined in the sum (36), after the matrix $P$ is sparsified by eliminating very small values below a threshold. We found that a threshold of $1/\sqrt{n}$ is sufficient for an acceptable approximation, resulting in a typical 90% sparsity pattern.

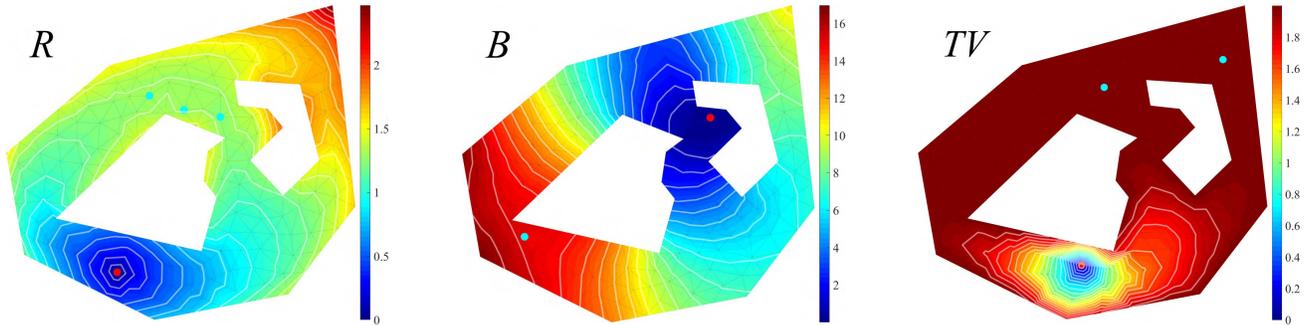

Fig. 4. Convex Holes shape at low resolution ($n = 175$). Spurious local minima for **(left)** resistance $R$ **(center)** biharmonic $B$ and **(right)** $TV$ distances on a low-resolution triangle mesh of Fig. 3. The red point is the target and the cyan points are local minima.

**Low resolution discretizations**
Obviously the higher the resolution of the triangulation, the closer the discrete setting approaches the continuous case, for which we can prove the equivalence of our distance functions $TV$ and $KL$ to the Dirichlet Green's function $D$. For low resolution triangulations, it is not obvious that a similar equivalence result holds, even if the discrete functions in principle have properties analogous to their continuous counterparts. The main issue is spurious local minima.

The discrete Green's functions $D$ and $N$, as the heat kernels $HD$ and $HN$, have maximum principles, so will never contain spurious local minima. It is not so obvious what happens with the resistance distance $R$, the biharmonic distance $B$, and the $TV$ and $KL$ divergence distances. It seems that $R, B$ and $TV$ all suffer from this affliction when the resolution of the discretization is too low. See Fig. 4. We did not encounter spurious minima at all in any of our experiments with $KL$.

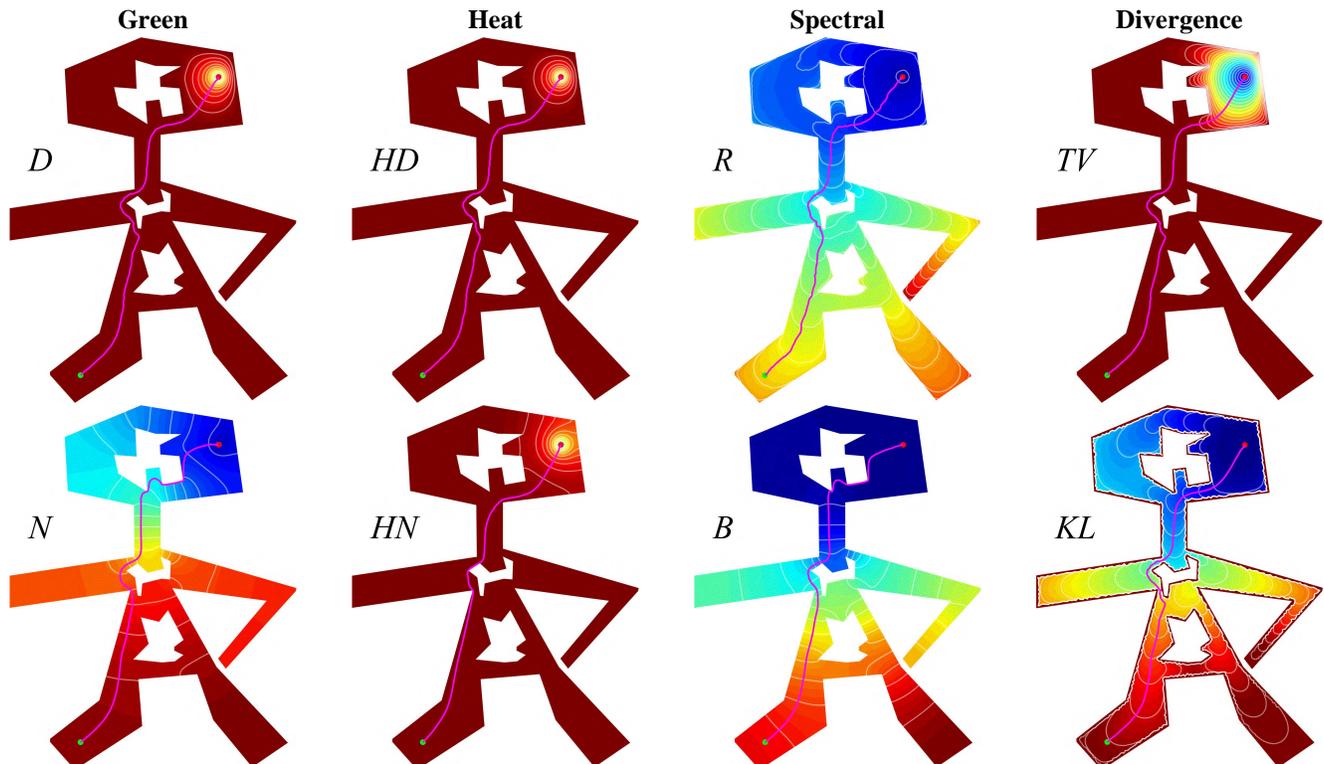

Fig. 5. HiRes Man2 shape ($n = 7,117$).

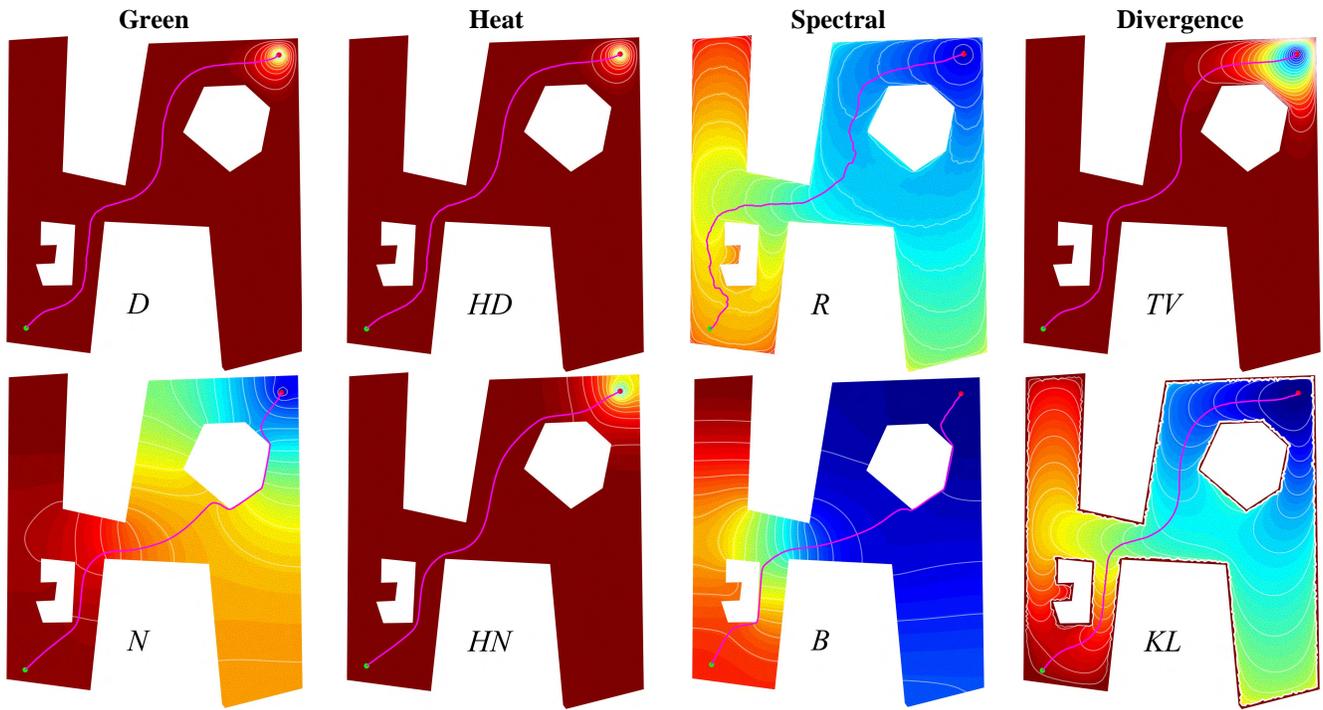

Fig. 6. Concave Holes shape ($n = 6{,}646$).

**More Examples**

Figures 5 and 6 compare the different paths generated on a high resolution discretization of some interesting 2D shapes. The results are quite consistent with what we observed on other domains: The equivalence of $D, TV$ and $KL$, the very flat character of $D, HD, HN$ and $TV$, and the noisiness of $R$.

Maze-type domains, as explored in the robotics community, are especially challenging inputs. This is because they are typically "one-dimensional" or "long and skinny" simply-connected domains, and the paths generated are quite long. Fig. 7 shows our results on the maze featured in Fig. 2 of Wray et al. [29]. On these types of domains, the fact that any interior point $q$ is quite close to some boundary point has two interesting and quite significant consequences: 1) $KL(q,p) \approx \log D(q,p)$, 2) Most of the Poisson kernel vector $P(p)$ is negligible, thus may be approximated well as a sparse vector, significantly saving time in the computation of $KL$. As mentioned above, for these types of examples, it is easy to run into precision problems when computing some of the distance functions. The heat functions $H$ decay so rapidly that they are effectively zero at any precision, yielding no gradient. The Dirichlet Green's function $D$ also decays very rapidly but the floating-point mechanism is still able

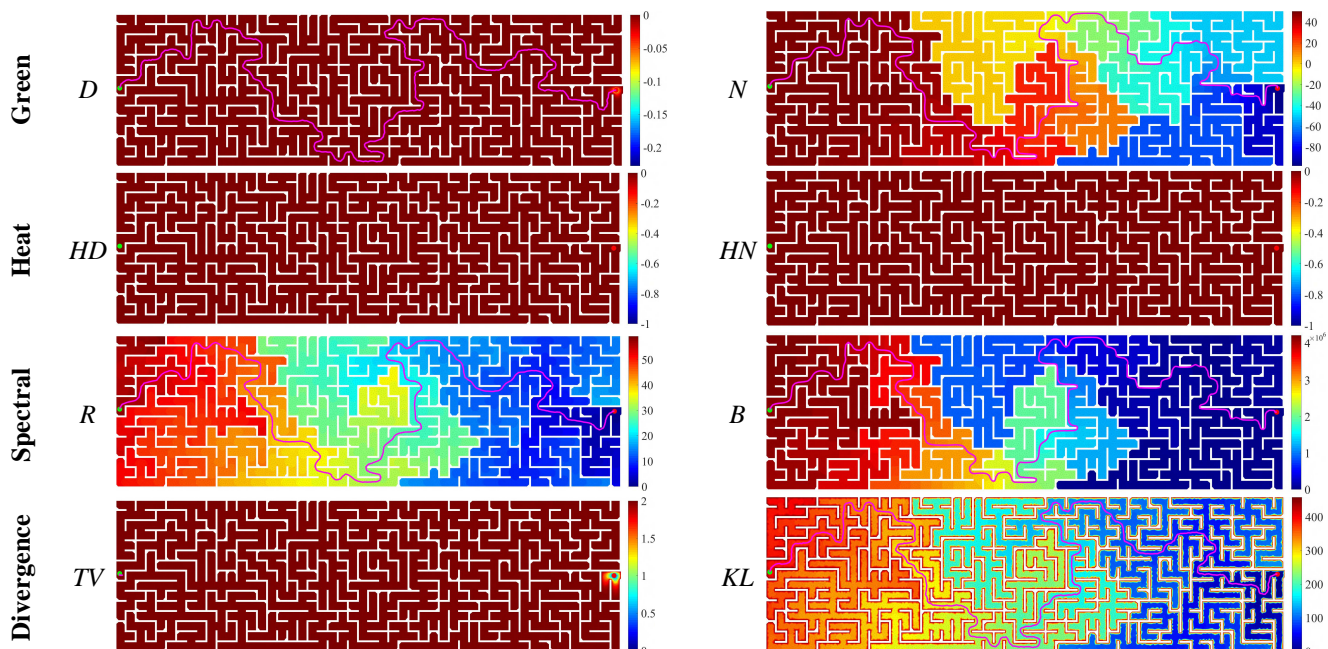

Fig. 7. Maze shape ($n = 18{,}929$).

## VI. CONCLUSION AND OPEN QUESTIONS

We have introduced a new family of planar distance functions based on the concept of $f$-divergences [8] which may be used to generate gradient-descent paths useful in robotic path planning. These distance functions make use of the concept of Poisson kernels (or harmonic measure) $P(t, z)$ on the domain boundary:

$$d_f(z, w) = \oint_{\partial \Omega} P(t, z) f\left(\frac{P(t, w)}{P(t, z)}\right) dt$$

where $f$ is a convex function such that $f(1) = 0$. We show that all $f$ distance functions based on $f$-divergences are equivalent to the Dirichlet Green's function, i.e. generate exactly the same gradient-descent paths, but may be computed much more efficiently. We have focused on two specific members of the family which are sufficiently different and representative: $TV$ (Total Variance) is the simple $L_1$ distance between two Poisson kernel vectors and $KL$ is the more sophisticated Kullback-Leibler divergence between two probability vectors. A natural generalization of the $TV$ distance is to use

$$f(x) = |1 - x|^p$$

for integer $p \geq 1$. The special case $p = 1$ coincides with $TV$, and the special case $p = 2$ with the well-known Chi-Squared distance. A natural generalization of the $KL$ distance ($f(x) = -\log x$ or $f(x) = x \log x$) is the $\alpha$-divergence for $\alpha \neq \pm 1$:

$$f(x) = \frac{4}{1 - \alpha^2}\left(1 - x^{\frac{1+\alpha}{2}}\right)$$

There are many more examples of interesting and popular $f$-divergences in the literature, e.g. the Helliger divergence:

$$f(x) = \left(\sqrt{x} - 1\right)^2$$

It remains to be seen which $f$-divergence is optimal for path-planning, where a major concern is its sensitivity to discretization.

Our Equivalence Theorem (Theorem A6 in the Appendix) has been proven for the case of a simply connected domain using the conformal invariance property. The analogous proof for the more general case of multiply connected domains seems to be more complicated and alludes us presently. However, we are confident, based on our experimental results, that the claim is true for this case too.

We have proved the Equivalence Theorem for the continuous case, and it has been experimentally observed to be true also when the domain is discretized by a dense triangulation and a discrete Laplacian used instead of a continuous one. We have detected a number of cases where spurious minima emerge in low resolutions. We speculate that there exists a condition on the minimal triangulation resolution such that the claim holds also for the discrete case above that resolution. This is an interesting topic for future research.

All of the theory and experimentation in this paper has been on planar domains. However, there exists a conformal theory (the so-called uniformization theorems [20]) also between manifold surfaces in three dimensions. Thus we speculate that results similar to ours could be obtained for path-planning on these types of surfaces, in particular terrains.

Finally, we note that another application of the conformal invariant distance functions could be in shape matching, as explored by Bronstein and Bronstein [4].

## APPENDIX

## CONFORMALLY INVARIANT DISTANCE FUNCTIONS

### 1. Conformal Mappings and Invariance

A distance function on a domain $\Omega$ is a non-negative real function $d: \Omega \times \Omega \to \mathbb{R}$ such that $\forall z \in \Omega, \ d(z, z) = 0$. In this Appendix we focus on *conformally invariant* distance functions in a simply connected domain $\Omega$, namely a distance function $d^\Omega$ such that

$$d^\Omega(p, q) = d^{C(\Omega)}\big(C(p), C(q)\big)$$

where $p, q \in \Omega$ and $C: \Omega \to \Omega'$ is a conformal mapping. In a nutshell, we say that a complex-valued function $C$ is *conformal* iff it is *holomorphic* (sometimes called *analytic*), namely a function of $z$ alone (and not of $\bar{z}$), and, in addition, its derivative never vanishes in $\Omega$. Conformal mappings are a cornerstone of complex analysis and have many important properties. We refer the interested reader to the classic textbook by Nehari [20] on this subject. In general the superscript of $d$ makes explicit which domain is referred to. When this is obvious, the superscript may be omitted.

**The Green's function**

A simple and well-known example of a conformally invariant function is $D$ – the Green's function of $\Omega$ with Dirichlet boundary conditions mentioned in Section 2:

**Theorem A1 [20, 11]:** $D$ is a conformal invariant, namely, if $C$ is a conformal mapping on $\Omega$ and $p, q \in \Omega$, then

$$D^\Omega(p, q) = D^{C(\Omega)}\big(C(p), C(q)\big)$$

**QED**

Although not strictly a distance function, $D$ is used as such for path planning based on gradient-descent.

In the special case where the domain is the unit disk, there is another well-known distance function which is a conformal invariant – the *hyperbolic* distance. This distance is actually a metric:

$$HB(w, z) = 2 \tanh^{-1}\left(\frac{|w - z|}{|1 - \bar{w}z|}\right) \qquad (A1)$$

**The divergence distances**

Let $\Omega$ be a general simply connected domain and let $P^\Omega(t, z)$ be the Poisson kernel of $\Omega$, where $t$ is a point on $\partial \Omega$, namely, for each $t \in \partial \Omega$, $P^\Omega(t, z) = P_t^\Omega(z)$ is the solution to the following Laplace equation with Dirichlet boundary conditions:

$$\nabla P_t^\Omega = 0$$
$$P_t^\Omega|_{\partial \Omega} = \delta(t)$$

$P^\Omega(t,z)$ is also known as the density of the *harmonic measure* [11] of $\Omega$, namely, for a subset $E \subset \partial\Omega$ of the boundary, the harmonic measure of $E$ relative to $z$ is:
$$\omega^\Omega(E,z) = \int_E P^\Omega(t,z)dt$$
$\omega^\Omega(E,z)$ is well known to be equal to the probability (density) that a random walk starting at $z$ will first exit $\Omega$ through $E$. As expected, by definition:
$$\omega^\Omega(\partial\Omega, z) = \int_{\partial\Omega} P^\Omega(t,z)dt = 1, \quad \forall z \in \Omega$$
Harmonic measure is well known to be, like the Green's function, a conformal invariant. Namely if $C$ is a conformal map on $\Omega$, then
$$\omega^\Omega(E,z) = \omega^{C(\Omega)}(C(E), C(z))$$
We focus now on the family of *divergence* distance functions, namely, those based on applying *f*-divergences [8] to the harmonic measure:
$$DV^\Omega(p,q) = \oint_{\partial\Omega} P^\Omega(t,p) f\left(\frac{P^\Omega(t,q)}{P^\Omega(t,p)}\right) dt$$
where $f: \mathbb{R}^+ \to \mathbb{R}$ is a strictly convex real function such that $f(1) = 0$.

The *KL* distance function is obtained as the special case $f(x) = -\log(x)$ and the *TV* distance function as the special case $f(x) = |1 - x|$.

**Theorem A2:** $DV$ is a conformal invariant, namely, if $C$ is a conformal mapping on $\Omega$ and $p, q \in \Omega$, then
$$DV^\Omega(p,q) = DV^{C(\Omega)}(C(p), C(q))$$

**Proof:** The relationship between the Green's function and the Poisson kernel is through the normal derivative at the boundary [1]:
$$P^\Omega(t,z) = \frac{\partial D}{\partial n}(t,z)$$
So, applying Theorem A1 and invoking the chain rule for a holomorphic function
$$P^\Omega(t,z) = \frac{\partial D^\Omega}{\partial n}(t,z) = \frac{\partial D^{C(\Omega)}}{\partial n}(C(t), C(z))|C'(t)|$$
$$= P^{C(\Omega)}(C(t), C(z))|C'(t)|$$
where $C(t)$ is the function obtained by restriction $C: \partial\Omega \to \partial C(\Omega)$. Consequently
$$DV^\Omega(p,q) = \oint_{\partial\Omega} P^\Omega(t,p) f\left(\frac{P^\Omega(t,q)}{P^\Omega(t,p)}\right)dt$$
$$= \oint_{\partial C(\Omega)} P^{C(\Omega)}(C(t), C(p))|C'(t)|f\left(\frac{P^{C(\Omega)}(C(t),C(q))|C'(t)|}{P^{C(\Omega)}(C(t),C(p))|C'(t)|}\right)\frac{dC(t)}{|C'(t)|}$$
$$= \oint_{\partial C(\Omega)} P^{C(\Omega)}(C(t), C(p)) f\left(\frac{P^{C(\Omega)}(C(t),C(q))}{P^{C(\Omega)}(C(t),C(p))}\right) dC(t)$$
$$= DV^{C(\Omega)}(C(p), C(q))$$
**QED**

We are mostly interested in the behavior of the *gradient* of the divergence distance. We start by stating a simple relationship between the gradients of a conformally invariant distance function on two conformally related domains:

**Theorem A3:** If $C$ is a conformal mapping on $\Omega$ and $d$ is a conformally invariant distance function (from a fixed point $p \in \Omega$) on $\Omega$, then
$$\nabla d^\Omega(q) = \nabla d^{C(\Omega)}(C(q))\frac{\overline{\partial C}}{\partial z}(q)$$
**Proof:** Recall that the gradient of $d$ as a complex number is $\nabla d = 2\frac{\overline{\partial c}}{\partial z}$, and apply the chain rule for a holomorphic function.
**QED**

A gradient-descent path between two points is generated by following the (direction of the) gradient of the distance function, which is possible iff that gradient does not vanish. We now show that, if it exists, the gradient-descent path of a conformally invariant distance function $d$ is also a conformal invariant:

**Theorem A4:** Denote by $G_d^\Omega(p,q)$ the gradient-descent path between $p$ and $q$ in a simply-connected domain $\Omega$, generated by a conformally invariant distance function $d$. If $C$ is a conformal mapping on $\Omega$ and $p, q \in \Omega$, then:
$$C\left(G_d^\Omega(p,q)\right) = G_d^{C(\Omega)}(C(p), C(q))$$
**Proof:** Let $z \in G_d^\Omega(p,q)$. By definition, the tangent to $G_d^\Omega(p,q)$ at $z$ is $-\nabla d^\Omega(z)$. Consider the point $C(z) \in C\left(G_d^\Omega(p,q)\right) \subset C(\Omega)$. The tangent there is $-\nabla d^\Omega(z)/\frac{\overline{\partial c}}{\partial z}(z)$, which, by Theorem A3, is just $-\nabla d^{C(\Omega)}(C(q))$.
**QED**

2. **Some Special Cases**

2.1 **The unit disk and $p = 0$**

We may analyze the behavior of the divergence distance functions and obtain exact expressions for these and other distance functions and their gradients when $\Omega = \mathbb{D}$ (the unit disk). We start with the canonical case that $p = 0$, and call this domain $\mathbb{D}^0$.

**The Green's function**
It is well known that
$$D^{\mathbb{D}_0}(z,0) = -\frac{1}{2\pi}\log|z| \qquad (A2)$$
and
$$\nabla D^{\mathbb{D}_0}(z) = r_D(|z|)z, \qquad r_D(x) = \frac{1}{2\pi x^2} \qquad (A3)$$
This indicates that the gradient of $D$ on the unit disk, when the target is the origin, is radial and never vanishes (for $z \neq 0$).

**The hyperbolic distance**
Based on (A1), we conclude that on $\mathbb{D}^0$:
$$HB^{\mathbb{D}_0}(z,0) = 2\tanh^{-1}|z| = \log\left(\frac{1+|z|}{1-|z|}\right) \qquad (A4)$$
and
$$\nabla HB^{\mathbb{D}_0}(z) = r_{HB}(|z|)z, \qquad r_{HB}(x) = \frac{2}{x(1-x^2)} \qquad (A5)$$

So *HB* has behavior similar to *D*. Note that although the distance has a global minimum of 0 at $z = 0$, the gradient is undefined there since the distance is not differentiable.

**The divergence distances**

The following theorem proves that general divergence distances on $D^0$ behave similarly to the Green's function *D* and the hyperbolic distance *HB*.

**Theorem A5:** If *DV* is a divergence distance based on $f$, then
$$DV^{D_0}(z, 0) = g(|z|)$$
and
$$\forall z \neq 0, \quad \nabla D^{D_0}(z, 0) \neq 0$$

**Proof:** The Poisson kernel on the unit disk D ($z \in D, w \in \partial D$) is:
$$P^D(w, z) = \frac{1}{2\pi} \frac{1 - |z|^2}{|w - z|^2}$$
It is easy to see that $P^D(w, 0) \equiv \frac{1}{2\pi}$, thus
$$DV^{D_0}(z, 0) = \frac{1}{2\pi} \int_0^{2\pi} f\left(\frac{1 - |z|^2}{|e^{i\theta} - z|^2}\right) d\theta$$
For brevity, we will drop the second argument of $DV^{D_0}$. Assume $z = |z|e^{i\alpha}$, and substitute $\phi = \theta - \alpha$:
$$DV^{D_0}(z) = \frac{1}{2\pi} \int_0^{2\pi} f\left(\frac{1 - |z|^2}{1 - |z|e^{i(\theta - \alpha)} - |z|e^{i(\alpha - \theta)} + |z|^2}\right) d\theta$$
$$= \frac{1}{2\pi} \int_{-\alpha}^{2\pi - \alpha} f\left(\frac{1 - |z|^2}{1 - |z|e^{i\phi} - |z|e^{-i\phi} + |z|^2}\right) d\phi =$$
$$= \frac{1}{2\pi} \int_0^{2\pi} f\left(\frac{1 - |z|^2}{1 - 2|z|\cos\phi + |z|^2}\right) d\phi = g(|z|)$$
The equality of the last two integrals is because both integrate the entire unit circle. Thus $DV^{D_0}(z)$ is a function of $|z|$ only.

For the gradient of $DV^{D_0}$, we will show that $\forall x \in (0,1)$, $g'(x) > 0$. Indeed, if
$$h(x, \phi) = \frac{1 - x^2}{1 - 2x\cos\phi + x^2}$$
then
$$g'(x) = \frac{1}{2\pi} \int_0^{2\pi} f'(h(x, \phi)) \frac{\partial h(x, \phi)}{\partial x} d\phi$$
Noticing that
$$H(x, \phi) = \int \frac{\partial h(x, \phi)}{\partial x} d\phi = \frac{2\sin\phi}{1 - 2x\cos\phi + x^2}$$
and integrating by parts:
$$g'(x) = \frac{1}{2\pi} f'(h(x, \phi)) H(x, \phi) \Big|_{\phi=0}^{\phi=2\pi}$$
$$- \frac{1}{2\pi} \int_0^{2\pi} f''(h(x, \phi)) \frac{\partial h(x, \phi)}{\partial \phi} H(x, \phi) d\phi$$
Since
$$H(x, 0) = H(x, 2\pi) = 0$$
and
$$\frac{\partial h(x, \phi)}{\partial \phi} = -\frac{2(1 - x^2)x \sin\phi}{1 - 2x\cos\phi + x^2}$$
we finally have

$$g'(x) = \frac{1}{2\pi} \int_0^{2\pi} f''(h(x, \phi)) \frac{4(1 - x^2)x \sin^2\phi}{(1 - 2x\cos\phi + x^2)^2} d\phi$$
For $x \in (0,1)$:
$$4(1 - x^2)x \sin^2\phi > 0$$
and
$$1 - 2x\cos\phi + x^2 \geq 1 - 2x + x^2 = (1 - x)^2 > 0$$
Since $f$ is strictly convex, $f'' > 0$, resulting in
$$g'(x) > 0$$
Going back to the gradient of $DV^{D_0}$:
$$\frac{\partial DV^{D_0}(z)}{\partial z} = \frac{g'(|z|)\sqrt{\bar{z}}}{2\sqrt{z}} = \frac{g'(|z|)\bar{z}}{2|z|}$$
and
$$\nabla DV^{D_0}(z) = 2\overline{\left(\frac{\partial DV^{D_0}}{\partial z}(z)\right)} = r_{DV}(|z|)z,$$
$$r_{DV}(x) = \frac{g'(x)}{x}$$
Since $g'(x) > 0$ in $(0,1)$, we conclude that $\nabla DV^{D_0}(z) \neq 0$ for $|z| \in (0,1)$.

**QED**

For the special case $KL^{D_0}$ we have $f(x) = -\log x$, thus
$$KL^{D_0}(z) = \frac{1}{2\pi} \int_0^{2\pi} -\log\left(\frac{1 - |z|^2}{1 - 2|z|\cos\phi + |z|^2}\right) d\phi$$
$$= -\log(1 - |z|^2) \tag{A6}$$
Thus
$$\nabla KL^{D_0}(z) = r_{KL}(|z|)z, \quad r_{KL}(x) = \frac{2}{1 - x^2} \tag{A7}$$
For the special case $TV^{D_0}$ we should be more careful, because $f(x) = |1 - x|$ is not strictly convex everywhere and not differentiable at 0. Thus the general derivation of Theorem A5 is not applicable and we must resort to an explicit computation:
$$TV^{D_0}(z) = \frac{1}{2\pi} \int_0^{2\pi} \left|1 - \frac{1 - |z|^2}{|e^{i\theta} - z|^2}\right| d\theta$$
$$= -\frac{1}{2\pi i} \int_D \left|1 - \frac{1 - |z|^2}{|w - z|^2}\right| \frac{dw}{w}$$
The integrand changes sign at two points $w_1$ and $w_2$ on the unit circle. Geometrically these are the two points where a circle through 0 and $q$ is tangent to the unit circle. By solving a quadratic equation, these points are:
$$w_{1,2} = \frac{z}{|z|}\left(|z| \pm i\sqrt{1 - |z|^2}\right)$$
We note that $\frac{w_1}{z} = \overline{\left(\frac{w_2}{z}\right)}$, leading (after much tedious algebra) to:
$$TV^{D_0}(z) = \frac{1}{\pi i} \int_{w_1}^{w_2} \left(1 - \frac{1 - |z|^2}{w|w - z|^2}\right) dw$$
$$= 4 + \frac{4i}{\pi}\left(\log\left(-\sqrt{1 - |z|^2} + i|z|\right)\right)$$
$$= \frac{4}{\pi} \sin^{-1}|z| \tag{A8}$$
with derivative:
$$\frac{\partial TV^{D_0}(z)}{\partial z} = \frac{2}{\pi |z|\sqrt{1 - |z|^2}} \bar{z}$$
leading to:

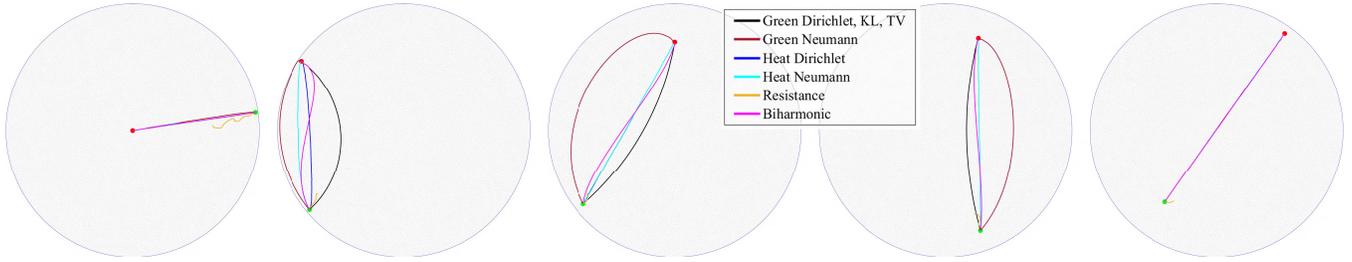

Fig. A1. Gradient-descent path generated by different distance functions between pairs of points on the unit disk. Green point is source and red is target.

$$\nabla TV^{D_0}(z) = r_{L1}(|z|)z, \quad r_{TV}(x) = \frac{4}{\pi}\left(\frac{1}{x\sqrt{1-x^2}}\right) \quad (A9)$$

As with the Poincaré hyperbolic metric, $TV$ has a global minimum at 0 but its gradient is undefined there.

**Comparison**

To summarize, these are the relevant entities for the distance functions mentioned above.

| $d$ | $f(x)$ | $g(x)$ | $r(x)$ |
|---|---|---|---|
| $D$ | – | $\dfrac{\log x}{2\pi}$ | $\dfrac{1}{2\pi x^2}$ |
| $HB$ | – | $2\tanh^{-1} x$ | $\dfrac{2}{x(1-x^2)}$ |
| $DV$ | strictly convex | $\dfrac{1}{2\pi}\displaystyle\int_0^{2\pi} f\left(\dfrac{1-x^2}{1-2x\cos\phi+x^2}\right)d\phi$ | $\dfrac{g'(x)}{x}$ |
| $KL$ | $-\log x$ | $-\log(1-x^2)$ | $\dfrac{2}{1-x^2}$ |
| $\chi^2$ | $x^2-1$ | $\dfrac{1+x^2}{1-x^2}$ | $\dfrac{4}{(1-x^2)^2}$ |
| $TV$ | $|1-x|$ | $\dfrac{4\sin^{-1} x}{\pi}$ | $\dfrac{4}{\pi x\sqrt{1-x^2}}$ |

Note that since all the $r$ functions are positive real functions, the negative gradient of the distance functions always points towards $p = 0$, albeit with different magnitudes. This means that the contours of the distances to $p = 0$ are concentric circles around 0 and the gradient-descent paths are always a straight line towards $p$. Note also that (a) for $TV$ and $KL$, the gradient grows infinitely as $q$ approaches the boundary and (b) when $z$ is close to the boundary, namely $|z|$ is close to 1, we have $\log D(z) \approx KL(z)$. Most importantly, all the distance functions but $D$ monotonically increase from 0 on (0,1), so have a global minimum at $p = 0$ and no local minima in the unit disk. $D$ monotonically increases from $-\infty$.

### 2.2 The unit disk and $p \neq 0$

The more general case where $p \neq 0$ can also be analyzed relatively easily thanks to the conformal invariance of the distance functions. This gives us an explicit form for gradient-descent paths between any two points in the unit disk.

The Riemann Mapping Theorem [20] guarantees that there exists a conformal mapping $C: \Omega \to D^0$ from any simply connected domain $\Omega$ to the unit disk such that $C(\partial\Omega) = \partial D^0$ and $C(p) = 0$. In general this conformal mapping cannot be obtained in closed form, but for the unit disk it is known to be the Mobius transformation $M: D \to D^0$: one that maps the unit disk to itself such that $p$ is mapped to 0:

$$C(z) = \frac{z-p}{1-\bar{p}z}$$
$$\frac{\partial C(z)}{\partial z} = \frac{1-|p|^2}{(1-\bar{p}z)^2} \quad (A10)$$

implying, by Theorems A1 and A2:

$$\nabla d^D(q) = \nabla d^{D_0}(C(q))\overline{\frac{\partial C(q)}{\partial q}} = r_d(|C(q)|)C(q)\frac{1-|p|^2}{(1-p\bar{q})^2}$$
$$= r_d(|C(q)|)\frac{1-|p|^2}{|1-\bar{p}q|^2}\left(\frac{q-p}{1-p\bar{q}}\right) \quad (A11)$$

So, here again, all distance functions have a unique minimum at $q = p$, as expected.

**Gradient-descent paths**

Let us now determine the common gradient-descent paths of all the distance functions between two points $q$ and $p$ in the unit disk D. In the case $p = 0$ (the special domain $D^0$), Theorem A5 has shown that the paths are just the straight (radial) lines between $q$ and the disk center. Theorem A4 implies that these are mapped through the same Moebius transformation to the general case of $p \neq 0$. Since circles (including straight lines, which are circles of infinite radius) are invariant to Moebius transformations, we conclude that all gradient-descent paths in the disk are circles. It is easy to see that the path between $p$ and $q$ is the *hyperbolic geodesic* between $q$ and $p$ in the Poincare hyperbolic disk model, namely the arc of the circle through $p$ and $q$ which is orthogonal to the unit circle. This will be a straight line only if $p$ and $q$ are on a diameter of the unit disk. Moreover, if $q$ is close to the boundary, the gradient will always be radial, independent of $p$.

Figure A1 shows examples of the gradient-descent paths generated by a number of distance functions studied in this paper, in the unit disk. Note the following: (1) The resistance distance $R$ path is very noisy and gets stuck very quickly. (2) All other paths are straight lines to the target when it is the origin. (3) The Dirichlet Green's function $D$ (and the equivalent $TV$ and $KL$) paths are hyperbolic circles. (4) The Neumann Green's function $N$ path is attracted to the boundary. (5) The biharmonic $B$ path is "wavy". (6) The heat paths $HD$ and $HN$ are very similar when distant from the boundary.

## 3. Equivalence of the distance functions

We have explicitly shown that the distance functions $D, HB$ and all variants of $DV$ are *equivalent* on the unit disk D, namely generate identical gradient-descent paths. This was due to the fact that they are equivalent on $D^0$ (all generate radial paths) and that carries over to D because of the conformal invariance of the distance functions and their gradient-descent paths. The same logic shows the equivalence of these distance functions on *all* simply connected domains.

**Theorem A6 (Equivalence Theorem):** If $DV$ is a divergence distance function, then $D, HB$ and $DV$ are equivalent on any simply connected domain $\Omega$.

**Proof:** In Section 2.1, including in Theorem A5, we have shown that any of these distance functions $d$ generate valid gradient-descent paths on $D^0$, i.e. its gradient never vanishes in $D^0$ except at the target, namely, for any $q \in \Omega$:
$$q \neq 0 \Longrightarrow \nabla d^{D^0}(q) \neq 0 \qquad (A12)$$
We now show the same for $\Omega$. Indeed, let $C: \Omega \to D^0$ be the conformal map (guaranteed to exist by the Riemann Mapping Theorem) that maps $\partial C$ to $\partial D^0$ and $C(p) = 0$. Theorem A3 implies
$$\nabla d^{\Omega}(q) = \nabla d^{D_0}(C(q)) \frac{\overline{\partial C}}{\partial z}(q)$$
Since $q \neq p$ implies $C(q) \neq 0$, and the conformality of $C$ implies $\frac{\overline{\partial C}}{\partial z}(q) \neq 0$, combining with (A12) implies
$$q \neq p \Longrightarrow \nabla d^{\Omega}(q) = \nabla d^{D_0}(C(q)) \frac{\overline{\partial C}}{\partial z}(q) \neq 0$$
Now consider this valid gradient-descent path generated by $d$ between $p$ and $q$ in $\Omega$. By (A2), (A4) and Theorem A5, for all the $d$'s, the path between $C(q)$ and $C(p) = 0$ in $D^0$ is the straight (radial) line segment between $C(q)$ and 0. Theorem A4 then implies, due to the conformal invariance of both distance functions, that all $d$'s generate the same path between $p$ and $q \in \Omega$. Hence all the distance functions are equivalent in $\Omega$.
**QED**